\documentclass[prl,twocolumn,showpacs,showkeys]{revtex4}%
\usepackage{amsfonts}
\usepackage[T2A]{fontenc}
\usepackage[cp1251]{inputenc}
\usepackage{amsmath}
\usepackage{amssymb}
\usepackage[russian,english]{babel}
\usepackage{graphicx}

\begin{document}

\title{On the possibility of making the complete computer model of a human
brain}

\author{A.V. Paraskevov}

\thanks{E-mail address: paraskev@kiae.ru}

\affiliation{Russian Research Center "Kurchatov Institute",
Kurchatov Sq. 1, Moscow 123182, Russia}

\begin{abstract}

The development of the algorithm of a neural network building by
the corresponding parts of a DNA code is discussed.

\end{abstract}



\maketitle

The aim of this short note is to point out to the concrete
opportunity of building of a complete computer model of the human
brain. At present time an architecture of the artificial neural
networks (ANNs), as well as the properties of neurons forming
them, is set by their developers empirically. The difficulty lies
in that the brain has a huge number of neurons and yet a lot more
number of the connections between them. Besides, in itself a
neuron is a very complex cell which functions as an element of
information processing are not yet studied completely. There is no
even a hint at the theory that can answer the question about an
order of the formation of the connections between neurons at an
embryo development. (This question pretends to the generality
since the structure of a cerebral cortex of infants is almost
identical.) That is why a computer simulation of the biological
neuronets is embarrassed even in a view of the essential progress
in the field of an experimental diagnostics of a brain. On the
other hand, the requirements in the ANNs which are commensurable
by the efficiency with a human brain grow. The progress in
molecular biology and genetics enables to build some algorithm,
which work results in a generation of the \textit{full-featured
}computer model of a brain. The point is that the nature has
created a compact memo book which contains all the data about the
structure and operation principles of a brain. This book is a
deoxyrybonucleic acid (DNA) molecule. DNA of an organism consists
of a set of genes which is called a genome. The genes are the
instructions (for some molecular machines) of how to create the
cells and tissues of an organism. For example, it is known that
the number of genes involved in development and functioning of a
human brain equals 3195 (see, e.g., \cite{1}). Thus the creation
of the brain model is reduced to the decoding corresponding parts
of a DNA code and an iterative construction of an ANN in
accordance with the "instructions" obtained. Certainly, the
realization of such an algorithm will demand an enormous computing
capacity and computer memory. However, the fast growth of these
characteristics gives a hope for such an opportunity.

Note that the character of an interaction of the model with a
suitable biological sample (it can be, for example, an alive part
of human body) can serve as a simple criterion of the model
operability. If the sample and the model communicate with each
other as two biological samples among themselves (or as two parts
of one sample), the model is evidently adequate and complete.

\end{document}